\documentclass{llncs} %
\usepackage[T1]{fontenc}
\usepackage{amsmath,amsfonts,bm}

\def\eqref#1{equation~\ref{#1}}
\def\1{\bm{1}}

\DeclareMathAlphabet{\mathsfit}{\encodingdefault}{\sfdefault}{m}{sl}
\SetMathAlphabet{\mathsfit}{bold}{\encodingdefault}{\sfdefault}{bx}{n}

\newif\ifcommentsoff
\commentsofftrue  %
\makeatletter
\newcommand{\authorcomment}[3]{%
  \ifcommentsoff
    \@bsphack\@esphack%
  \else
    \@bsphack
    \textcolor{#2}{[#1: #3]}%
    \@esphack
  \fi
}
\makeatother

\newcommand{\dimitra}[1]{\authorcomment{Dimitra}{cyan}{#1}}

\newcommand{\dc}[1]{\authorcomment{DC}{red}{#1}}
\newcommand{\nf}[1]{\authorcomment{NF}{blue}{#1}}

\newcommand{\approach}[0]{\arc}
\newcommand{\soundness}[0]{soundness\xspace}
\newcommand{\arcexpanded}[0]{\textsc{Automated Reasoning checks~(ARc)}\xspace}
\newcommand{\arc}[0]{\textsc{ARc}\xspace}

\newcommand{\pmcexpanded}[0]{\textsc{Policy Model Creator~(PMC)}\xspace}
\newcommand{\pmc}[0]{\textsc{PMC}\xspace}
\newcommand{\avexpanded}[0]{\textsc{Answer Verifier (AV)}\xspace}
\newcommand{\av}[0]{\textsc{AV}\xspace}
\newcommand{\policymodel}{policy model\xspace}

\newcommand{\policymodels}{policy models\xspace}

\newcommand{\NOTRANSLATION}{\textit{NoTranslations}\xspace}
\newcommand{\VALID}{\textit{Valid}\xspace}
\newcommand{\SATISFIABLE}{\textit{Satisfiable}\xspace}
\newcommand{\INVALID}{\textit{Invalid}\xspace}
\newcommand{\AMBIGUOUS}{\textit{TranslationAmbiguous}\xspace}
\newcommand{\IMPOSSIBLE}{\textit{Impossible}\xspace}
\newcommand{\TOOCOMPLEX}{\textit{TooComplex}\xspace}
\newcommand{\cqa}{\textsc{ConditionalQA-logic}\xspace}

\newcommand{\figurem}[1]{Fig.~\ref{#1}}

\usepackage{stmaryrd}

\newcommand{\smtlib}[1]{{\small\texttt{#1}}}

\newcommand{\myparagraph}[1]{\noindent\emph{#1}}

\usepackage{amssymb}
\usepackage{booktabs}
\usepackage{enumitem}
\usepackage{graphicx}
\usepackage{hyperref}
\usepackage{lipsum}
\usepackage{multirow}
\usepackage{subcaption}
\usepackage{tabularx}
\usepackage[table,dvipsnames]{xcolor}
\usepackage[most]{tcolorbox}
\usepackage{url}
\usepackage{xspace}
\usepackage{marvosym}
\usepackage{algorithm}
\usepackage[noend]{algpseudocode}
\usepackage{amsmath,amssymb,amsfonts}
\usepackage{mathtools}
\usepackage{mathpartir}
\usepackage{thmtools}
\usepackage{wrapfig}
\usepackage{soul} %

\usepackage{tcolorbox}
\tcbuselibrary{breakable}
\definecolor{codegreen}{rgb}{0,0.6,0}
\definecolor{codegray}{rgb}{0.5,0.5,0.5}
\definecolor{codepurple}{rgb}{0.58,0,0.82}
\definecolor{codered}{rgb}{0.6,0,0}
\definecolor{backcolour}{rgb}{0.95,0.95,0.92}
\definecolor{backcolour2}{rgb}{0.95,0.92,0.95}
\lstdefinelanguage{SMTLIB}{
    keywords={
        set-logic, set-info, set-option, declare-sort, define-sort,
        declare-fun, define-fun, declare-const, define-const,
        assert, check-sat, get-model, get-value, get-proof,
        get-unsat-core, exit, push, pop, reset,
        and, or, not, ite, let, forall, exists, distinct,
        Int, Real, Bool, String, Array, BitVec,
        +, -, *, div, mod, abs, =, <, <=, >, >=,
        true, false,
        select, store,
        str.len, str.++, str.substr, str.contains, str.indexof,
        concat, extract, bvnot, bvand, bvor, bvxor,
        if, then,
    },
    keywordstyle=\color{blue}\bfseries,
    keywords=[2]{
        sat, unsat, unknown, model
    },
    keywordstyle=[2]\color{magenta}\bfseries,
    comment=[l]{;},
    commentstyle=\color{codegreen}\itshape,
    string=[b]",
    stringstyle=\color{codepurple},
    basicstyle=\ttfamily\tiny,
    backgroundcolor=\color{backcolour},
    numbers=none,
    numberstyle=\tiny\color{codegray},
    breaklines=true,
    breakatwhitespace=false,
    keepspaces=true,
    showspaces=false,
    showstringspaces=false,
    showtabs=false,
    tabsize=2,
    frame=single,
    framerule=0.5pt,
    rulecolor=\color{gray!50}
}

\usepackage{tikz}
\usetikzlibrary{positioning, shapes.geometric, arrows.meta, fit, calc}

\setlist{nolistsep}
\setlist[itemize]{leftmargin=*,noitemsep, topsep=0pt}

\title{
A Neurosymbolic Approach to Natural \\ 
Language Formalization and Verification
}

\author{
\textbf{Chenyang~An\textsuperscript{1}},
\textbf{Sam~Bayless\textsuperscript{1}}, 
\textbf{Stefano~Buliani\textsuperscript{1}}, 
\textbf{Darion~Cassel\textsuperscript{1}},
\textbf{Byron~Cook\textsuperscript{1,2}}, 
\textbf{Duncan~Clough\textsuperscript{1}}, 
\textbf{Rémi~Delmas\textsuperscript{1}},
\textbf{Nafi~Diallo\textsuperscript{1}}, 
\textbf{Ferhat~Erata\textsuperscript{1}}, 
\textbf{Nick~Feng\textsuperscript{1}}, 
\textbf{Dimitra~Giannakopoulou\textsuperscript{1}}, 
\textbf{Aman~Goel\textsuperscript{1}}, 
\textbf{Aditya~Gokhale\textsuperscript{1}},
\textbf{Joe~Hendrix\textsuperscript{1}},
\textbf{Victor~Heorhiadi\textsuperscript{1}},
\textbf{Marc~Hudak\textsuperscript{1}}, 
\textbf{Dejan~Jovanović\textsuperscript{1}}, 
\textbf{Andrew~M.~Kent\textsuperscript{1}}, 
\textbf{Benjamin~Kiesl-Reiter\textsuperscript{1}(\Letter)},
\textbf{Jeffrey~J.~Kuna\textsuperscript{1}}, 
\textbf{Nadia~Labai\textsuperscript{1}}, 
\textbf{Joseph~Lilien\textsuperscript{1}}, 
\textbf{Divya~Raghunathan\textsuperscript{1}}, 
\textbf{Zvonimir~Rakamari\'c\textsuperscript{1}},
\textbf{Niloofar~Razavi\textsuperscript{1}}, 
\textbf{Michael~Tautschnig\textsuperscript{1,4}}, 
\textbf{Ali~Torkamani\textsuperscript{1}}, 
\textbf{Nathaniel~Weir\textsuperscript{1}}, 
\textbf{Michael~W.~Whalen\textsuperscript{1}},
\textbf{Jianan~Yao\textsuperscript{3}}
}
\institute{
\textsuperscript{1}\,Amazon Web Services
\textsuperscript{2}\,University College London
\textsuperscript{3}\,University of Toronto
\textsuperscript{4}\,Queen~Mary~University~of~London
}

\begin{document}

\maketitle
\begin{abstract}

Large Language Models perform well at natural language interpretation and reasoning, but 
% their inherent stochasticity 
their lack of formal correctness guarantees 
limits their adoption in regulated industries like finance and healthcare that operate under strict policies.
To address this limitation, we launched \arcexpanded: a public service that (1) uses LLMs with optional human guidance to formalize natural language policies, allowing fine-grained control of the formalization process, and (2) uses inference-time autoformalization to validate logical correctness of natural language statements against those policies.
\approach performs multiple redundant formalization steps at inference time, checking the formalizations for semantic equivalence. 
% Our benchmarks show that \approach exceeds 99\% \soundness, a near-zero false positive rate in identifying logical validity.
Our benchmarks show that \approach exceeds 99\% \soundness and achieves a near-zero false positive rate in identifying logical validity.
Our approach produces auditable artifacts that substantiate the verification outcomes and can be used to improve the original text. 
% NOTE: legal/PR message here, do not tweak
\approach is the first commercial offering from a major cloud provider to integrate automated reasoning into a generative AI guardrail.
% END legal/PR
\end{abstract}

\section{Introduction}
\label{sec:introduction}

The capabilities of Large Language Models (LLMs) continue to advance rapidly, demonstrating unprecedented improvements in coherence and analytical accuracy~\cite{wei2022chain,yao2023tree,lewis2021retrievalaugmentedgenerationknowledgeintensivenlp}. Despite these advances, their 
tendency to generate plausible but incorrect information (hallucinations, cf.~\cite{DBLP:journals/corr/abs-2401-11817}) remains a barrier to widespread adoption in regulated sectors. Industries such as healthcare, financial services, and legal practices have legal and regulatory obligations for accuracy and auditability that current LLM technology has yet to meet~\cite{DBLP:journals/npjdm/HaltaufderheideR24}.

Companies develop institutional policies to ensure compliance with
laws and regulations. 
Such policies are typically captured in natural language (NL), defining rules, procedures, or guidelines.
When deploying LLMs to answer questions about these policies, a key challenge emerges: can we develop \emph{guardrails} to ensure that the LLM outputs are correct?
Consider an airline implementing a chatbot to assist customer service representatives in navigating refund policies: if the chatbot incorrectly claims that a customer is eligible for a refund,
this could lead to legal exposure and loss of customer trust. 

An effective guardrail would enable representatives to rely on chatbot responses by ensuring that when it reports an answer as valid, it actually is.
Inspired by the concept of soundness in logic, we define \emph{soundness} in our context as $(1-r)$, where $r$ is the rate of incorrect validity claims across all decisions. High soundness thus means that across all requests, incorrect approvals are rare. Following established practices in safety-critical systems, where reliability is often measured in ``nines'' (e.g., 99\% = ``two nines,'' 99.9\% = ``three nines''), we target soundness levels of at least 99\%, and secondarily focus on recall to maximize the probability of accepting valid content.
We also pursue actionable feedback that steers LLMs toward content that a conservative guardrail can accept.

A natural candidate for robust 
guardrails are symbolic reasoning systems, as they
use
formal logic to generate independently verifiable guarantees \cite{robinson2001handbookofar}. This 
aligns well with policy documents, which rely on logical, rule-like statements (e.g., ``if a flight is canceled or \ldots, then passengers are entitled to a refund''). But symbolic methods struggle with interpreting natural language, 
triggering
the development of neurosymbolic approaches 
that combine the NL processing capabilities of neural networks with the mathematical rigor of symbolic systems~\cite{hitzler21neurosymbolic}.

This paper presents \arcexpanded, a neurosymbolic approach exceeding 99\% \soundness on datasets it was not trained on. High soundness is also reflected in metrics such as false positive rate and precision. 
In addition, \arc delivers explainable verdicts and provides actionable feedback for revising LLM outputs. 
% NOTE: legal/PR message here, do not tweak
\approach is the first commercial offering from a major cloud provider to integrate automated reasoning into a generative AI guardrail.
% END legal/PR.

\arc operates through two complementary components. The first, called \pmcexpanded, combines LLMs with symbolic reasoning to translate NL policies into formal \emph{policy models} expressed in logic. 
It
begins with an autoformalization phase that generates an initial policy model. 
This is 
followed by an optional vetting phase where domain experts 
refine the policy model with assistance from the system. 
Vetting enables domain experts to resolve ambiguities and inconsistencies
in the original documents, 
or to correct potential imprecisions from autoformalization. Policy model creation occurs offline; its computation cost will be amortized across subsequent verification tasks. 

The second component, called \avexpanded, 
verifies NL content against \policymodels. 
The \av uses LLMs to translate NL content into individual logical claims in the scope of the \policymodel. Each claim is analyzed separately and assigned a verification result, together with detailed logical explanations and corrective guidance.
To increase reliability, the \av uses multiple LLMs to simultaneously formalize the same NL content, then uses symbolic reasoning to compare formalizations and assign confidence scores.
The \av delivers auditable logical artifacts that substantiate the verification outcomes. 

\section{Related Work}
\label{sec:related}
Recent approaches use LLMs as judges to evaluate factual accuracy~\cite{jacovi2025facts}, though these rely on 
% the same probabilistic models that introduce errors. 
the same LLMs that are subject to inaccuracies to evaluate LLM-generated content.
MiniCheck~\cite{tang-etal-2024-minicheck} provides efficient fact-checking by decomposing claims. RefChecker~\cite{hu-etal-2024-knowledge} introduces knowledge-centric verification against structured knowledge bases. 
SelfCheckGPT~\cite{manakul-etal-2023-selfcheckgpt} leverages consistency across responses to detect hallucinations. FactCheck-GPT~\cite{wang-etal-2024-factcheck} provides comprehensive evaluation with fine-grained error categorization.
While promising, these methods 
% are probabilistic in nature and 
cannot provide formal guarantees. Our neurosymbolic framework verifies logical validity against formalized policies, achieving near-zero false positives.

Neurosymbolic systems combine LLMs (sometimes enhanced with Chain-of-Thought prompting~\cite{xu-etal-2024-faithful,liu-etal-2025-logic,Xiong2024HYBRIDMINDMS}) with symbolic reasoning. They typically translate NL to formal representations that are then solved by external reasoners~\cite{pan-etal-2023-logic,olausson-etal-2023-linc,callewaert-etal-2025-verus,ryu2024divide}.
LINC~\cite{olausson-etal-2023-linc} uses first-order logic with Prover9~\cite{mccune2005release}.
Verus-LM~\cite{callewaert-etal-2025-verus} provides a multi-paradigm framework with IDP-Z3~\cite{carbonnelle2022interactive}. 
SAT-LM~\cite{ye-etal-2023-satlm} employs declarative prompting with SMT~\cite{z3-solver}. Logic-LM~\cite{pan-etal-2023-logic} supports multiple formalisms with self-refinement. 
Other approaches use custom DSLs~\cite{ganguly2024proofthoughtneurosymbolic} or Answer Set Programming~\cite{ishay2023leveraging,yang2023coupling,brewka2011answer}.

These neurosymbolic systems formalize both the background knowledge and the query in a single step for each problem instance (or, in the case of~\cite{yang2023coupling}, rely on hand-written background knowledge). In contrast, \approach autoformalizes both but separates them: the policy model can be vetted independently (through linting, symbolic test case generation, and human review) before being reused across many verifications. Regarding translation correctness, most of these systems offer only syntactic error detection and basic consistency checks (e.g., satisfiability of the formalized theory). LINC goes further by sampling multiple translations from a single model and applying majority voting over solver outcomes (Valid, Invalid, Unknown) rather than comparing translations at the logic level. \approach cross-checks per-query translations across diverse models using a notion of symbolic equivalence, with confidence thresholds indicating semantic agreement.

Autoformalization has been studied extensively in mathematics~\cite{wang2018first,szegedy2020promising,wu2022autoformalization,jiang2022draft}. In contrast to \approach, these efforts typically assume that the background knowledge (the mathematical theory) has been pre-formalized, often by humans, and only the problem to be solved must be formalized by an LLM.

There is also a substantial body of work on dedicated formal languages for capturing policies and legislation, see, e.g., the Programming Languages and the Law (ProLaLa) workshops at POPL. Catala~\cite{merigoux2021catala} is a notable example: a domain-specific language designed for systematic translation of statutory law into executable code, using defeasible reasoning to handle the general-case/exceptions logic common in legal texts. Such approaches capture laws directly in the dedicated language, whereas \approach autoformalizes existing NL policy documents into logic. Casadio et al.~\cite{casadio2026nlp} propose a methodology for certifying the robustness of NLP models, showing how verification errors can stem from the NL representation layer. Their work on robustness guarantees for NL models is complementary to our redundant translation approach for mitigating translation errors.

\section{Methodology}
\label{sec:overview}

\begin{figure}[t]
  \centering
  \includegraphics[width=1.0\textwidth,alt={Pipeline diagram with two main components. Top: Policy Model Creator (PMC) takes a natural-language policy document, splits it into text spans, autoformalizes each span with self-refinement, and composes them into a policy model. It then applies logical test case enumeration and conflict detection, which serve as user feedback that can be leveraged for manual and LLM-guided repair to arrive at the final policy model. Bottom: Answer Verifier (AV) takes a question-answer conversation, produces multiple autoformalizations, aggregates them via logical equivalence, verifies against the policy model using an SMT solver, and outputs verification results with feedback.}]{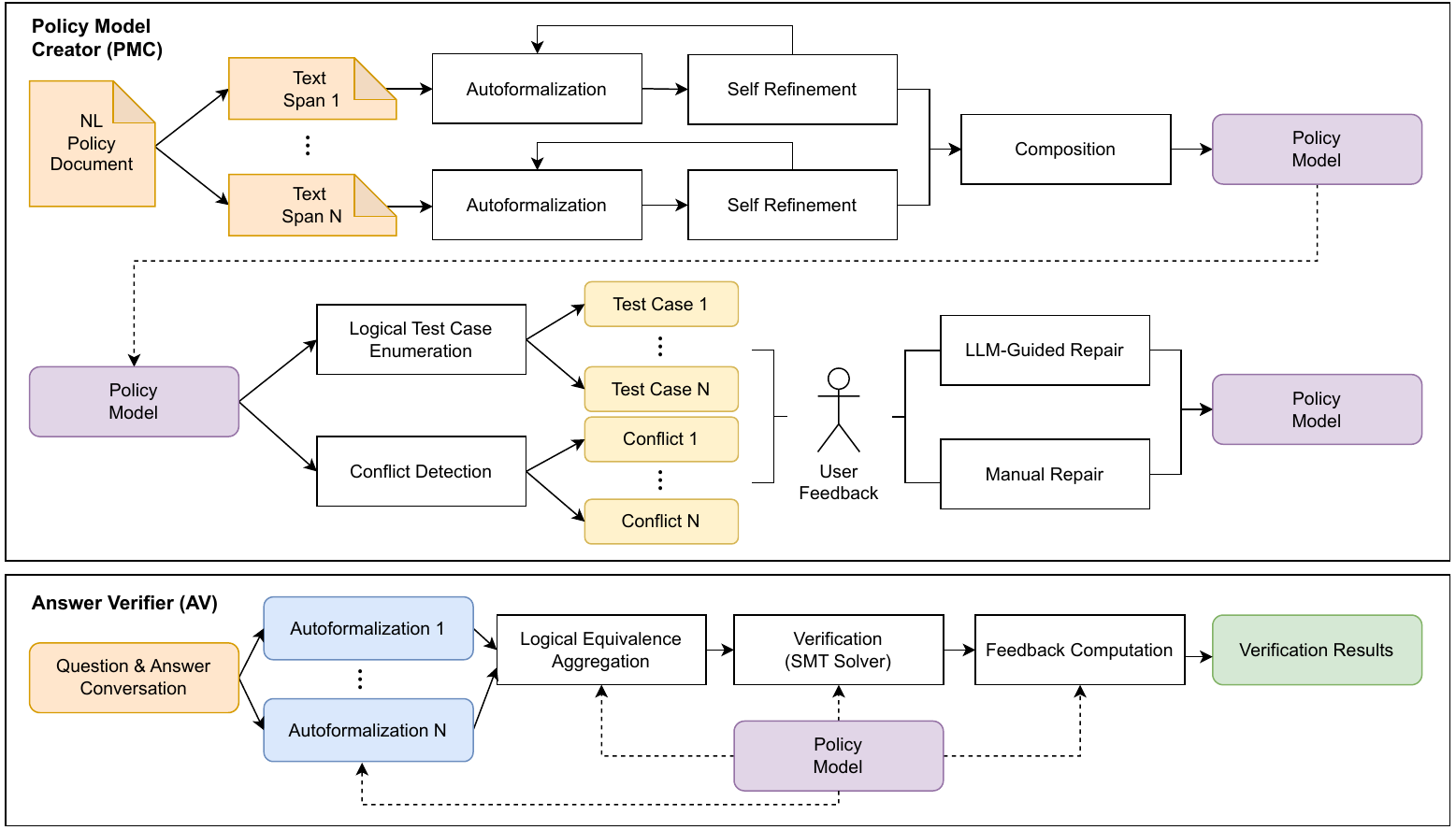}
  \caption{End-to-end architecture of \arc}
  \label{fig:architecture-overview}
\end{figure}

\figurem{fig:architecture-overview} shows \arc's two main components: the \pmc (\S\ref{sec:method_dmc}) and the  \av (\S\ref{sec:method_av}). We illustrate our approach with an NL policy about park admission fees:

\vspace{3pt}
\noindent\fbox{\begin{minipage}{0.98\textwidth}
    \footnotesize
    \centering
    \vspace{5pt}
    \begin{minipage}{0.97\textwidth}
        \textbf{General admission}: The regular admission to the park is \$50. The admission fee in the low season is 75\% of the regular admission fee. \\[0.5em]
        \textbf{Discount}: Seniors (age greater than 65) qualify for a 40\% discount. Whenever a discount applies, there will be a \$10 flat discount processing fee. 
        \\[0.3em]
        \textbf{Credit}: You can use credit for up to 50\% of your final admission  (1 credit for 1 dollar). However, if credits are used, then the discount rate is capped at 25\%. You can purchase credit at a rate of \$0.60 per credit. You can only purchase credit in increments of 5 (cost \$3).
        \\[0.3em]
        \textbf{Tax}: A federal tax of 10\% applies to the final expense.
    \end{minipage}
    \vspace{2pt}
\end{minipage}}

\vspace{5pt}
Suppose a user asks ``I am a senior and want to visit the park in the low season, and I have a budget of \$35.40. Can I visit the park?'', and we want to verify a chatbot's answer of ``No, \$35.40 is not enough.''

\approach tackles the verification problem in two stages.
In the first stage, the \pmc autoformalizes the policy into a \emph{\policymodel}: a set of logic rules, expressed in SMT-LIB~\cite{BarFT-SMTLIB}, together with a \emph{schema} that defines variables with their types and NL descriptions.
SMT-LIB is a standardized logic language that uses prefix notation, where operators precede their arguments; e.g., ``if $x$, then $y$'' is written as \smtlib{(=> x y)}.  %
\figurem{fig:policymodel} shows snippets of the \policymodel.
In the second stage, the \av autoformalizes the statement under validation into logic formulas (over the \policymodel schema), then uses an SMT solver to verify those formulas against the \policymodel. The result includes the logic translation (with a confidence score between 0 and 1), the validation result, 
%and feedback explaining the result (snippet shown in~\figurem{fig:socratesfeedback}, details in \S\ref{sec:method_av}).
and further feedback (see~\figurem{fig:socratesfeedback}).

In our example, the result is \textit{Satisfiable}:
%, meaning the statement is not always valid: 
the claim (cannot visit the park) is consistent with the premises (the person is a senior, it is low season, and they have a fund of \$35.40 available) but doesn't necessarily follow from them. \av provides two assignments as feedback: 
one showing a counter-example where admission is %in fact 
possible, and a case where the person %indeed 
cannot enter the park. 
The key difference is in the use of credits ($\mathit{creditUnit = 3}$ vs. $\mathit{creditUnit = 0}$). 

\myparagraph{Explanation.} 
%In order to be admitted, the person needs to use \$15 worth of credits. 
The person needs to use \$15 worth of credits for admission. 
The admission fee in the low season is \$37.5 (75\% of \$50). After applying the 25\% senior discount (capped at 25\% because of credit use) and adding the \$10 discount processing fee, the actual admission fee becomes \$$38.125$. 
This fee can be paid by combining \$15 of credit (cost: \$9) with a \$$23.125$ cash payment, for an expense of \$$32.125$. 
After adding a federal tax of 10\%, the final expense becomes \$$35.3375$, which is within the budget of \$35.40. 
%Notably, this problem proved challenging even for advanced LLM judges: 
Both Claude Sonnet 3.7 and Opus 4.1 (with reasoning mode) incorrectly classified the answer as valid, providing plausible but flawed reasoning (see Appendix Figs \ref{fig:llmjudge_sonnet37} and \ref{fig:llmjudge_opus41}).

% (a counter-example to validity, shown in \S\ref{sec:method_av}). 

\begin{figure}[tb]
\small
    \begin{subfigure}[b]{0.99\textwidth}
        \renewcommand{\arraystretch}{1.2} 
        \begin{tabularx}{\linewidth}{|l|c|X|}
        \hline
        \textbf{Variable} & \textbf{Type} & \textbf{Description} \\
        \hline
        isLowSeason & Bool & Whether the admission day is in the low season\\
        feeAfterDiscount & $\mathbb{R}$ & 	
        Admission fee after discounts are applied but before tax \ \\
        \hline
        \end{tabularx}
        \label{fig:policymodelvariable}
    \end{subfigure}
    
    \begin{subfigure}[b]{\textwidth}
        \fbox{\begin{minipage}{0.97\textwidth}
            \centering
            \vspace{0.5em}
            \begin{minipage}{0.96\textwidth}
            \textbf{Rule 1:} 
            \smtlib{(={}> isLowSeason (= admissionFee (* 0.75 baseFee)))}\\[0.5em]
            \textbf{Rule 2:} 
            \smtlib{(<{=} (* 2.0 customerCredits)  feeAfterDiscount)}
            
            \end{minipage}
            \vspace{0.5em}
        \end{minipage}}
        \label{fig:policymodelrules}
    \end{subfigure}
    \caption{Snippets of \policymodel (top: variable schema; bottom: rules)}
    \label{fig:policymodel}
\end{figure}

\begin{figure}[bt]
\small
    \fbox{\begin{minipage}{0.98\textwidth}
        \vspace{0.5em}
        \begin{minipage}{0.97\textwidth}
            \textbf{Translation (Confidence: 1.0)} \\[0.2em]
            $P$: \smtlib{(and (= ageClass SENIOR) isLowSeason (= totalAdmissionFund 35.4))} \\
            $C$: \smtlib{(not isEntryAllowed)}
            
            \textbf{Result:} \SATISFIABLE (not \VALID)
            
            \textbf{Counter-Example} (shows $C$ can be false): \\
            \hspace*{1em}\begin{tabular}[t]{@{}l@{}}
            creditUnit=3, customerCredits=15.0, creditDollarValue=9.0, \\
            cashAmount=23.181, totalPaymentAvailable=38.181, \\
            finalAdmissionFee=38.125, isEntryAllowed=true, \ldots
            \end{tabular}
            
            \textbf{Satisfying Assignment} (shows $C$ can be true): \\
            \hspace*{1em}creditUnit=0, finalAdmissionFee=35.75, isEntryAllowed=false, \ldots
        \end{minipage}
        \vspace{0.5em}
    \end{minipage}}
    \caption{Snippet of validation feedback}
    \label{fig:socratesfeedback}
\end{figure}

\subsection{\pmcexpanded}
\label{sec:method_dmc}

The \pmc takes a policy document written in natural language and autoformalizes it into a \policymodel. It also provides an array of utilities to support users in policy vetting, described in detail in the \emph{Policy Vetting} paragraph below.
\subsubsection{Autoformalizing Policies}
% \label{sec:method_dmc:autoformalization}
%

To handle the size and complexity of real-world policy documents in the face of known LLM reasoning limitations around context size and distractors~\cite{rajeev2025,levy2024}, the \pmc takes a divide-and-conquer approach to autoformalize documents into logic (see \figurem{fig:architecture-overview}).

The \pmc first splits the input document into a set of text spans. These are processed using an incremental, refinement-guided autoformalization procedure: A language model processes each span and identifies statements that express coherent, formalizable meaning.
For each statement, the LLM translates the semantic content into a list of SMT-LIB datatypes, variables, and logical constraints~(\emph{rules}). The LLM's context maintains existing declarations within a span to avoid duplicated or conflicting declarations.
The complete formalization of a span is what we call a \emph{policy unit}. If this process introduces an error (e.g., malformed syntax), we provide the invalid declarations and their failure causes to the LLM for repair in a refinement loop.

After the \pmc has formalized all text spans, it composes the policy units into a single \policymodel.
The \pmc generates textual embeddings  of variables and clusters them using cosine similarity. Variables within a cluster are unified, while variables that share the same name but are not clustered are renamed.
Consistent replacement of original variables with unified variables is performed for the rules of each policy unit, then the rules are aggregated, dropping syntactic duplicates.
The resulting \policymodel is a structured representation of the document, consisting of three fields: datatypes, variables, and rules. Each variable is associated with an NL description that explains its meaning in terms of the source document, as shown in \figurem{fig:policymodel}.
This initial \policymodel is then vetted, as described in \S\ref{app:policy_vetting}. 
We measure the relationship between document size and formalized policy size in \S\ref{app:dmc-scaling}.

\subsubsection{\pmc Policy Vetting}
\label{app:policy_vetting}
The initial policy model might contain errors and omissions. Additionally, as shown in \S\ref{sec:refining_policy_study}, NL policies often contain ambiguities that only subject matter experts can resolve. 
We therefore provide users with several methods for vetting of their policy models: linting, inspection, and testing (both manual and automatic). We also develop automated repair approaches around these vetting methods.

\myparagraph{Linting.}
A linter for our \policymodels that checks integrity and consistency properties beyond the simple malformedness errors caught during autoformalization. We perform a mix of syntax-based and semantic checks: we detect unused variables and types, contradictory rules, and disjoint rule sets. The list of warnings is shown to the user, who can address them directly (e.g., by deleting an unused variable) or through more detailed policy inspection and repair.

\myparagraph{Inspection.}
Manual inspection allows users to review the generated policy model, similar to code review in software development. We assist users by providing two views of the rules for inspection: SMT-LIB for experts and structured English for non-experts. Structured English is generated using templates (like ``if \ldots\ then \ldots'') to avoid potential hallucinations from LLMs. Users also can provide NL feedback on the policy, which triggers an automatic LLM-based policy repair step that adjusts the policy model. Manual inspection provides strong correctness guarantees when all rules are carefully reviewed, but it can be challenging with large numbers of complex rules that have intricate interactions. The \pmc therefore also provides testing as an additional policy vetting methodology.

\myparagraph{Testing.}
Testing provides a systematic way to validate policy models through examples. 
Similar to unit tests, test cases in the \pmc are either NL question-answer pairs or simple NL statements with their expected outcomes provided by the user (e.g., valid, invalid). Test cases can either be provided manually by users or generated automatically.
For manually provided test cases, the \pmc ``executes'' them by running the \av to compute the findings; a mismatch indicates an error in the policy model or in the \av translation.
The \pmc also offers automatic, symbolic test-case generation that leverages an SMT solver to systematically explore the state space of the policy model. Since such test cases are generated symbolically, each comes with its provably-correct actual finding, so a mismatch isolates the error to the policy model.
Either way, the supplied information (e.g., rules justifying the result) helps users diagnose and repair.

\subsection{\avexpanded}
\label{sec:method_av}
The AV uses LLMs to translate natural language (i.e., question-answer pairs) into a list of premise-conclusion pairs, where premises ($P$) and conclusions ($C$) are expressed in the \policymodel's 
schema. For example, "Since you have at least \$50, you can enter the park" translates 
to premise $(\geq \mathit{totalAdmissionFund} ~ 50)$ and conclusion $\mathit{isEntryAllowed}$ which represent the contextual facts and logical consequences, respectively.

To increase translation confidence, the AV \emph{redundantly translates} the NL statement using $k$ LLMs (Alg.~\ref{alg:redundant}), then compares the resulting premise-conclusion pairs semantically using an SMT solver to estimate a confidence score for each pair. Intuitively, the confidence score of a premise-conclusion pair $\langle P, C \rangle$ is the proportion of the $k$ translations that
non-vacuously entail the implication $P \Rightarrow C$ (i.e., entail it without rendering $P$ contradictory). For example, consider the text from \S\ref{sec:overview} and the policy model in \figurem{fig:policymodel}. Redundant translation with three LLMs produces three identical premise-conclusion pairs, thus producing the results in 
\figurem{fig:socratesfeedback} with confidence score $3/3$ (1.0). If one LLM instead produced a pair with a different conclusion (\smtlib{isEntryAllowed}), \av 
would return two distinct pairs: the original from \figurem{fig:socratesfeedback} with confidence $2/3$,
and the alternative with confidence $1/3$.

\begin{algorithm}[t!]
\caption{\av Redundant Translation}
\begin{algorithmic}[1]
\small
\Procedure{RedundantTranslation}{$msg$, $\textit{policy}$, $LLMs$}
    \State $F \gets \emptyset$; $\textit{Ts} \gets$ [\textit{Translate}($msg$, $\textit{policy}$, $llm$) $\mid$ $llm \in \textit{LLMs}$] \label{ln:translate}
    \For{each $\langle P, C \rangle$ in every $T \in Ts$} \label{ln:loophead}
        \State $F$.add($\langle P, C, \mathit{cf}\rangle$) where $\mathit{cf} = |\{T' \in \textit{Ts} \mid T' \models (P \Rightarrow C) \land T' \not\models \neg P\}| / |Ts|$ \label{ln:conf}
    \EndFor \label{ln:loopend}
    \State \textbf{return} $F$
\EndProcedure
\end{algorithmic}
\label{alg:redundant}
\end{algorithm}

%\dimitra{Too little space before here}
\myparagraph{Validation Feedback.}
\label{sec:av_feedback}
After translation, the \av uses an SMT solver to validate each claim $\langle P, C, \mathit{conf}\rangle$ against $\mathcal{M}$ while producing logically grounded feedback given the following precedence order:

\begin{center}
\begin{tabular}{ll}
\toprule
\textbf{Finding} & \textbf{Condition} \\
\midrule
\TOOCOMPLEX & Text or SMT-LIB translation exceeds token limits \\
\AMBIGUOUS & Confidence below threshold (default: $3/3$) \\
\NOTRANSLATION & LLM cannot translate text to \policymodel vocabulary \\
\IMPOSSIBLE & $\mathcal{M} \vDash \lnot P$ (premises contradict \policymodel) \\
\INVALID & $\mathcal{M} \land P \vDash \lnot C$ (conclusion is inconsistent) \\
\VALID & $\mathcal{M} \land P \vDash C$ (conclusion is entailed) \\
\SATISFIABLE & $\mathcal{M} \land P \nvDash C$ and $\mathcal{M} \land P \nvDash \lnot C$ (consistent, not entailed) \\
\bottomrule
\end{tabular}
\end{center}

For \IMPOSSIBLE, \INVALID, and \VALID findings, the feedback includes relevant rules from the \policymodel extracted from the SMT solver, providing sufficient information for independent 
verification via theorem prover.
For \SATISFIABLE findings, the feedback returns assignments ("scenarios") demonstrating how the answer can be correct or wrong, thus providing hints on how the premises could be extended to make the conclusion valid (see~\figurem{fig:socratesfeedback}, for example, where the use of credits is a differentiator between the scenarios).
For \AMBIGUOUS findings, the feedback presents two differing translations, together with an assignment that is satisfiable in one translation but not in the other. 
\NOTRANSLATION findings return the untranslatable text segments. 
Logic warnings are surfaced if premises or conclusions are always true or false irrespective of policy rules.

\section{Empirical Evaluation}
\label{sec:evaluation}

In order to understand \approach's effectiveness as a guardrail, and the impact of our design choices, we evaluate \approach, and more specifically the \av, around the following research questions (RQs): 

\begin{description}[topsep=5pt]
    \item[{\sc RQ1 (Reliability of Validating Logical Accuracy):}] 
    How reliably does \approach validate logical accuracy compared to alternative baselines?

    \item [{\sc RQ2 (Impact of Redundant Translation):}]
How does redundant translation~(\S\ref{sec:method_av}) impact \approach's performance?

    \item[{\sc RQ3 (Effectiveness of \approach{}'s feedback):}] 
    Is \approach's feedback effective in driving improvement of LLM outputs?
\end{description}

\myparagraph{Metrics.}
We frame logical accuracy detection as a binary classification problem: decide whether NL statements are \VALID or not. 

To target high-stakes applications, our primary objective is to eliminate false positives across the entire pipeline. In this context, rejecting borderline cases (\textit{not-valid}) is favorable: such outputs can either be refined by the answer-generating LLM using \approach feedback, or escalated to human experts. We thus define \emph{soundness} as $1 - \frac{\#\text{False Positives}}{\#\text{Samples}}$, measuring the rate at which incorrect approvals occur across all decisions. 

A natural alternative to soundness would be precision. The two metrics differ in perspective: precision estimates a posterior (``Given a request was classified as \VALID, how likely is that verdict to be correct?''), while soundness estimates a prior (``How likely is a request to receive an incorrect \VALID verdict?''). Based on customer feedback, we prioritized the prior perspective: \approach's customers are service operators deploying chatbots across thousands of daily interactions, where the relevant operational question is how likely any given request is to be incorrectly approved. 
Like any soundness notion, our soundness rewards abstention 
and is incomplete on its own, so we use recall to complement it.
As mentioned in \S\ref{sec:introduction}, we target soundness levels of at least 99\%. 

In addition to soundness, we use standard classification metrics (precision, recall, F1, accuracy), treating \VALID as the positive class and all others as negative. When comparing alternative methods, \VALID recall is used as a tie-breaker under the requirement of maintaining high soundness. 

\subsection{Evaluation of Logical Accuracy Validation}
\label{sec:evaluate_logical_accuracy}

\begin{table}[t!]
    \caption{Comparison of logical accuracy detection performance on \cqa~\cite{sun-etal-2022-conditionalqa}. The columns show soundness (S), false-positive rate (FPR), precision (Pr), recall (Re), F1 score (F1), accuracy (Ac), counts of true/false positives/negatives (TP/FP/TN/FN), and error count (Error\#). Approaches meeting soundness threshold are highlighted, as well as best and worst values for other metrics.
    }
    \label{tab:rq1_detection}
    \centering
    \small
    \resizebox{\textwidth}{!}{
    \begin{tabular}{l|c|cc|ccc|cccc|c}
    \toprule
    \multicolumn{1}{c|}{Method} & S $\uparrow$ & FPR $\downarrow$ & Pr $\uparrow$ & Re $\uparrow$ & F1 $\uparrow$ & Ac $\uparrow$ & TP $\uparrow$ & FP $\downarrow$ & TN $\uparrow$ & FN $\downarrow$ & Error\# $\downarrow$ \\
    \midrule
    \approach (\#3-ensemble, threshold=3/3) & \cellcolor{ForestGreen!40}99.4 & 1.8 & 94.4 & 14.9 & 25.8 & 42.4 & 169 & 10 & 548 & 962 & 5 \\
    \approach (\#3-ensemble, threshold=2/3) & \cellcolor{ForestGreen!40}99.2 & 2.5 & 93.2 & 16.9 & 28.6 & 43.5 & 191 & 14 & 544 & 940 & 9 \\
    \approach (without redundant translation) & 98.4 & 4.8 & 92.2 & 28.0 & 43.0 & 50.2 & 317 & 27 & 531 & 814 & 1 \\
%    \approach (\#3-ensemble, threshold=0) & 98.8 & 3.6 & 96.9 & 54.8 & 70.0 & 68.4 & 621 & 20 & 535 & 513 & 8 \\
    \midrule
    LLMaJ (\#3-ensemble, threshold=3/3) & 98.2 & 5.4 & 92.4 & 32.4 & 47.9 & 52.9 & 366 & 30 & 528 & 765 & -\\
    LLMaJ (\#3-ensemble, threshold=2/3) & 97.9 & 6.3 & 92.1 & 36.3 & 52.0 & 55.2 & 410 & 35 & 523 & 721 & -\\
    LLMaJ (1x Sonnet4.5) & 97.9 & 6.3 & 92.0 & 35.4 & 51.1 & 54.6 & 400 & 35 & 523 & 731 & -\\
    LLMaJ (1x Sonnet4.5 w/ extended thinking) & 97.1 & 8.8 & 92.7 & 55.0 & 69.0 & 67.0 & 622 & 49 & 509 & 509 & -\\
    \midrule
    FG Implicit span-level~\cite{jacovi2025facts} & 95.0 & 15.2 & 90.4 & 70.6 & 79.2 & 75.3 & 798 & 85 & 473 & 333 & -\\
    FG JSON~\cite{jacovi2025facts} & 94.6 & 16.5 & 89.7 & 70.5 & 78.9 & 74.8 & 797 & 92 & 466 & 334 & -\\
    FG Response-level~\cite{jacovi2025facts} & 83.8 & 49.1 & 78.3 & \cellcolor{green!15}\textbf{87.6} & 82.7 & 75.5 & \cellcolor{green!15}\textbf{991} & 274 & 284 & \cellcolor{green!15}\textbf{140} & -\\
    MiniCheck~\cite{tang2024bespokeminicheck} & 88.5 & 34.9 & 83.0 & 84.4 & 83.7 & 78.0 & 954 & 195 & 363 & 177 & -\\
    RefChecker~\cite{hu-etal-2024-knowledge} & 91.9 & 24.6 & 86.3 & 76.0 & 80.8 & 75.8 & 860 & 137 & 421 & 271 & -\\
    SelfCheckGPT~\cite{manakul-etal-2023-selfcheckgpt} & 93.0 & 21.1 & 89.2 & 86.3 & \cellcolor{green!15}\textbf{87.7} & \cellcolor{green!15}\textbf{83.8} & 976 & 118 & 440 & 155 & -\\
    \midrule
    Logic-LM~\cite{pan-etal-2023-logic} & 98.0 & 5.9 & 86.4 & 18.6 & 30.6 & 43.5 & 210 & 33 & 525 & 921 & \cellcolor{orange!15}1149 \\
    Proof of Thought~\cite{ganguly2024proofthoughtneurosymbolic} & \cellcolor{orange!15}88.8 & \cellcolor{orange!15}33.9 & \cellcolor{orange!15}81.2 & 72.2 & 76.4 & 70.2 & 817 & \cellcolor{orange!15}189 & \cellcolor{orange!15}369 & 314 & \cellcolor{green!15}\textbf{0}\\
    LINC~\cite{olausson-etal-2023-linc} & \cellcolor{ForestGreen!40}\textbf{99.9} & \cellcolor{green!15}\textbf{0.2} & \cellcolor{green!15}\textbf{98.7} & \cellcolor{orange!15}6.6 & \cellcolor{orange!15}12.4 & \cellcolor{orange!15}37.4 & \cellcolor{orange!15}75 & \cellcolor{green!15}\textbf{1} & \cellcolor{green!15}\textbf{557} & \cellcolor{orange!15}1056 & 696\\
    \bottomrule
    \end{tabular}}
\end{table}

\myparagraph{Dataset.} 
Several 
reasoning benchmarks (e.g., FOLIO~\cite{han2024folio}, ProofWriter~\cite{tafjord2021proofwriter}, and LogicNLI~\cite{tian2021diagnosing}) test logical inference, but our focus is on 
validating whether NL answers comply with formalized policy documents.
We therefore focus on the ConditionalQA dataset~\cite{sun-etal-2022-conditionalqa} because it is well-aligned with our task, featuring: 1) questions that require compositional logical reasoning, 2) variety of questions (yes/no, multiple answers, not-answerable), and 3) human annotated answers.  

We enrich the ConditionalQA dev dataset (391 labeled QAs over 59 source documents)  with several types of ``not valid'' examples beyond its original ``valid'' / ``not\_answerable'' classification.
The resulting set includes the following categories: \VALID (logically correct), \INVALID (incorrect due to wrong conditions), \SATISFIABLE (missing necessary conditions), \IMPOSSIBLE (contradictory conditions), and \NOTRANSLATION (content that cannot be formalized, originally classified as not-answerable). 
These categories were created by systematically manipulating the conditional structure of original answers: removing conditions (\VALID $\rightarrow$ \SATISFIABLE), negating the claim (\VALID $\rightarrow$ \INVALID), or merging contradictory conditions (\VALID $\rightarrow$ \IMPOSSIBLE). The extended dataset (\cqa) contains 377 \VALID examples and 186 examples that are not \VALID (112 \INVALID, 56 \SATISFIABLE, 4 \IMPOSSIBLE, and 14 \NOTRANSLATION). \dimitra{Are we making these public? I forgot}

Our goal in this section is to evaluate the AV's soundness in an end-to-end setting with automatically generated, unvetted policy models. The policy models used for Table~\ref{tab:rq1_detection} were autoformalized by the \pmc and not manually refined. Section~\ref{sec:refining_policy_study} provides initial evidence that human vetting can further improve both soundness and recall.

\myparagraph{\textit{RQ1}: Reliability of Validating Logical Accuracy.}
Table~\ref{tab:rq1_detection} reports on the comparison of \approach against alternative methods: LLM-as-Judge (LLMaJ) with 
different prompting strategies, FACTS Grounding (FG)~\cite{jacovi2025facts}, fine-grained hallucination detection methods~\cite{tang-etal-2024-minicheck,tang2024bespokeminicheck,manakul-etal-2023-selfcheckgpt,hu-etal-2024-knowledge}, and other neurosymbolic approaches~\cite{pan-etal-2023-logic,ganguly2024proofthoughtneurosymbolic,olausson-etal-2023-linc}. This evaluation focuses on the ability of each approach to predict validation labels for QA pairs about given NL policy documents. We evaluate each of the 563 benchmark instances three times to account for nondeterminism in LLM outputs, yielding 1689 total decisions.

We first examine the approaches that meet our soundness threshold, which are only \approach and LINC~\cite{olausson-etal-2023-linc}. Between the two, \approach has a higher recall (14.9\% vs 6.6\%) and significantly fewer errors (5 vs 696 out of 1689 instances), where errors are instances in which the system was unable to formalize the QA contents and thus could not produce a verdict.
Second, we compare \approach to others that did not achieve the required soundness threshold of 99\%.
\approach's reliability comes with lower recall (14.9\% for the configuration with soundness over 99\%), but the tradeoff is intentional: in safety-critical domains, false approvals are far more costly than false rejections.
Among the methods below the soundness threshold, LLMaJ (\#3-ensemble, threshold=3/3) comes closest at 98.2\% soundness with precision similar to \approach's (92.4\% vs 94.4\%), but produces 3 times as many false positives (30 vs 10, or 5.4\% vs 1.8\% FPR).
In contrast, FG Response-level has the highest recall of 87.6\%, but this comes at the cost of soundness dropping to just 83.8\%, the second-lowest of all methods. \nf{TODO for CRC, if possible, we would like to cite why 99\% as a bar, since a lot of study depend on it}

\myparagraph{\textit{RQ2}: Impact of Redundant Translation.}
\approach uses redundant translation (Alg.~\ref{alg:redundant})
to increase confidence in NL-to-logic translations. Comparing the first 3 rows of Table~\ref{tab:rq1_detection}: soundness rises from 98.4\% to 99.4\% and FPR drops from 4.8\% to 1.8\%, at the cost of reduced recall (28.0\% to 14.9\%). Lowering the confidence threshold from $3/3$ to $2/3$ recovers recall to 16.9\%, with soundness dropping to 99.2\%.

\arc's recall is low compared to the other techniques in Table~\ref{tab:rq1_detection}, meaning that in many cases it returns results other than  \VALID for content that should be \VALID. \arc is intended to be deployed as a guardrail for chatbots in regulated industries, where an incorrect approval (e.g., `you qualify at 0\% interest') causes real harm at scale. In this setting, a non-\VALID verdict is not a failure: the system can revise the answer using \arc's feedback (see RQ3) or route to a customer support agent. Low recall thus means more deferrals, not more failures, and at 99.4\% soundness, \VALID verdicts carry strong enough assurance to reduce the review burden. In Section~\ref{sec:refining_policy_study}, we demonstrate that three rounds of revision with \approach's feedback can drive the rate of \VALID answers from 9\% to 46\% on a real-world policy.

\subsection{Refining Real-World Policy Models and Answers}
\label{sec:refining_policy_study}
\newlength{\validnfigwidth}\setlength{\validnfigwidth}{0.54\textwidth}
\begin{wrapfigure}{r}{\validnfigwidth}
    \includegraphics[width=\validnfigwidth,alt={Line chart showing the distribution of ARc validation findings across rounds 0 through 5 of answer revision. The x-axis is the revision round and the y-axis is the percentage of answers in each category. Valid findings rise steeply from about 9 percent at round 0 to 36 percent after one revision and 43 percent after two, reaching over 50 percent by round 5. Satisfiable findings drop sharply from about 35 percent to near 2 percent. Ambiguous stays relatively stable around 41 to 50 percent. Invalid and Impossible decrease to near 0 percent by round 5, while Unknown settles around 3 percent.}]{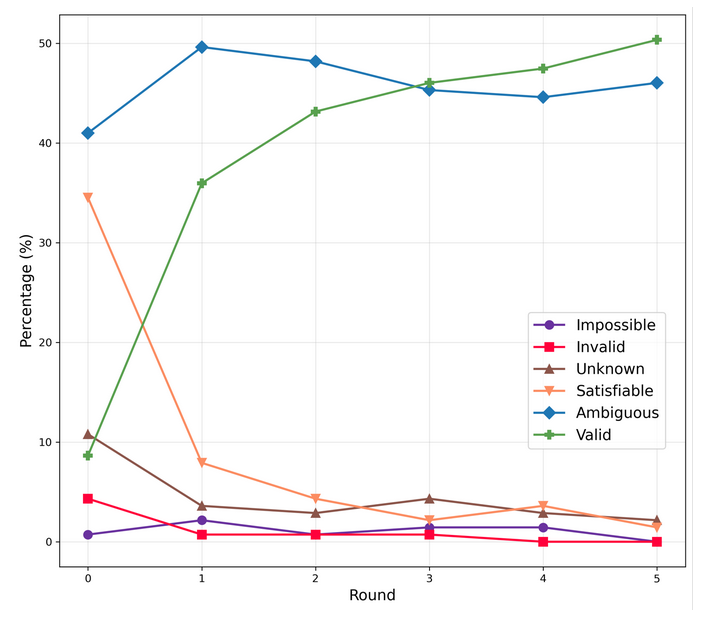} 
    \caption{\arc validation finding distribution after $k$ iterations of answer revision using \arc feedback. At $k=0$, we plot the finding distribution before any revisions.}
    \label{fig:curated-valid-n}
\end{wrapfigure}

To understand the applicability of \arc to real-world policies, we collected customer-facing policy documents from six different businesses, refined with human-in-the-loop vetting~(\S\ref{app:policy_vetting}).
We use these policies to evaluate iterated self-refinement of LLM answers using \arc feedback (\textit{RQ3}); in Appendix~\ref{app:experiments_policy_vetting}, we additionally share one policy as a case study of the manual refinement process itself (soundness: 96.8\%$\to$100\%, recall: 25\%$\to$45.5\%).

\myparagraph{\textit{RQ3}: Effectiveness of \approach{}'s Feedback.}
In a real-world setting, users may deploy LLM-based chatbots to answer questions about %these 
policies, requiring assurance of answer correctness.
The formally-grounded feedback that \arc provides (Section~\ref{sec:av_feedback}) can be used for automated answer revision: given a non-\VALID verdict, an LLM iteratively revises its answer using \arc's feedback.
\figurem{fig:curated-valid-n} shows how the relative percentages of each finding type evolve as an LLM iteratively revises \av-judged non-\VALID answers using \arc feedback (\#3-ensemble, threshold=3/3).
After
just three iterations, the LLM goes from 9\% to 46\% \VALID answers.
Primarily, this comes from a sharp reduction in \av-judged \SATISFIABLE answers (where the answer could be true or false depending on context not provided in the question or answer).
By providing logically-derived scenarios showing when the answer is true and when it is false, \arc enables the LLM to effectively revise these into \VALID answers.
\arc's feedback is less effective in revising \AMBIGUOUS and \NOTRANSLATION  answers. 
Our analysis shows that in these cases revising the \policymodel is more effective; e.g., the \policymodel could be missing variables, leading to a failure to formalize the answer, or the \policymodel could have variables that overlap in meaning, leading to a failure to generate a consistent formalization of the answer.
Average iterations to \VALID and success rate per finding type can be seen in Appendix Table~\ref{tab:avg-iterations-to-valid}.
\nf{TODO: Update Fig.4's legend to include NoTranslation}
\dc{It includes no translation}
\nf{TODO: It is hard to tell why we are less effective for \NOTRANSLATION from the plot}
\nf{TODO: I feel that we need to do a more detailed analysis on how cases status switch from round to round. The conclusion that we are very effective at revising Satisfiable cases could be misleading because they might be revised to be ambiguous (you see the jump of ambiguity between round 0 and 1, which correlates to the biggest drop of SAT). I found this conclusion a bit misleading based on my other experiments. }
\dc{Hopefully Table~\ref{tab:avg-iterations-to-valid} addresses these comments; it shows we're most effective at revising \SATISFIABLE $\to$ \VALID.}

\section{Conclusions, Limitations, and Future Work}
\label{sec:conclusions}
We presented \approach{}, a neurosymbolic guardrail that exceeds 99\% %(our ``two nines'' target) 
soundness when validating LLM answers against NL policies.
%, an assurance unattainable by existing approaches.  
This soundness comes at the cost of recall, a tradeoff we believe appropriate for regulated industries.
Since its launch, \approach has been adopted across industries, including responsible AI in education~\cite{pwc2024arc}, financial AI agents~\cite{icme2025arc}, and logistics operations~\cite{amazonlogistics2025arc}.
Soundness of \approach{} heavily depends on the quality of the \policymodel that it uses for validation. For this reason, \approach{} enables human oversight. 
% Such oversight is particularly relevant when policies are complex or very long documents. As we have shown, when domain experts refine policy models, both soundness and recall improve significantly. Beyond metrics, our formal representations let experts resolve ambiguities in policy documents, which is something no LLM can do, as only humans with authority can provide the definitive interpretation of what was intended.
As we have shown, \approach{} provides meaningful feedback to aid automated answer revision, resulting in an increase of \VALID answers. Despite \approach's high soundness, there are limitations:

%While \approach achieves strong soundness guarantees, limitations include:
%
\begin{itemize}
% \item \textit{Scalability}: Policy models from documents with hundreds of pages can include hundreds-to-thousands of rules, making them challenging for human vetting.
\item \textit{Document types}: Policies with numerical tables, cross-references, or implicit assumptions can be challenging to formalize without human vetting. % or background knowledge.
%\item \textit{Autoformalization challenges}: Subtle issues like ambiguous pronouns, implicit temporal or conditional scoping, and complex nested clauses can lead to incorrect formalizations.  This risk is only partially mitigated by redundant translations and human review.
\item \textit{Autoformalization challenges}: Subtle issues like ambiguous pronouns or implicit temporal scoping can lead to incorrect formalizations.
\item \textit{Computational cost}: Redundant translation requires multiple (3) LLM calls, resulting in average 5-15 second latency and increased API cost per Q/A validation with our current implementation.
%that propagate through the current pipeline.
\item \textit{Human effort}: The investment for human vetting of policies, while amortized over time, remains a significant upfront cost, especially for large documents. %with hundreds of rules. 
\item \textit{Dataset representativeness}: The mechanically mutated evaluation examples (removing conditions, negating claims, merging contradictory conditions) may not fully capture the kinds of errors real LLMs produce in practice. Evaluating on naturally occurring LLM failures is a direction for future work.
\end{itemize}
Future work includes exploring automatic and confidence-aware focused vetting, fine-tuned translation models for improving accuracy and latency/costs, and improved logical formalisms to address current limitations.
% while targeting three nines soundness and beyond.
Our approach directly benefits from advances in LLMs and generative AI techniques: as models improve, their ability to formalize natural language to logic will too. We are confident \approach will inherit these improvements while maintaining the mathematical guarantees provided by symbolic reasoning.

\begin{credits}
\subsubsection*{Disclosure of Interests.} The authors have no competing interests to declare that are relevant to the content of this article.
\end{credits}

\subsubsection*{Data-Availability Statement.} \approach is publicly available as part of Amazon Bedrock Guardrails at \url{https://aws.amazon.com/bedrock/guardrails/}. The evaluation uses a non-public augmentation of ConditionalQA~\cite{sun-etal-2022-conditionalqa}.

\bibliography{smaller}
\bibliographystyle{splncs04}

\vspace*{\fill} \pagebreak
\appendix

\section{Appendix}

\subsection{Additional Experiments}
\label{app:experiment_additional}
\subsubsection{Impact of Human Policy Vetting}
\label{app:experiments_policy_vetting}
We designed a case study to evaluate the support that \approach provides for policy refinement and its impact on \approach{}'s performance as a guardrail. 
In our study, we create and compare two formalizations of an airline's refund policy: one using the \pmc to create a \policymodel without additional vetting, and one that is revised using a human-in-the-loop (as described in \S\ref{app:policy_vetting}). 
To evaluate these two formalizations, we created a test suite balanced across three sources: 1) verbatim statements from the original policy; 2) Q/A pairs generated by an LLM; 3) Q/A pairs generated by three different individuals. The expected classification labels were determined manually.

\begin{table}[ht]
    \caption{Effect of human vetting on logical accuracy detection for RyanAir's customer service policy.}
    \label{tab:rq4_ryanair}
    \centering
    \small
    \resizebox{\textwidth}{!}{
    \begin{tabular}{l|cc|cccc|cccc}
    \toprule
    \multicolumn{1}{c|}{Method} & S $\uparrow$ & FPR $\downarrow$ & Pr $\uparrow$ & Re $\uparrow$ & F1 $\uparrow$ & Ac $\uparrow$ & TP $\uparrow$ & FP $\downarrow$ & TN $\uparrow$ & FN $\downarrow$ \\
    \midrule
    \approach with human vetting & \textbf{100.0} & \textbf{0.0} & \textbf{100.0} & \textbf{45.5} & \textbf{62.5} & \textbf{61.3} & \textbf{20} & \textbf{0} & \textbf{18} & \textbf{24} \\
    \approach without any human vetting & 96.8 & 8.7 & 84.6 & 25.0 & 38.6 & 43.6 & 11 & 2 & 16 & 33 \\
    \bottomrule
    \end{tabular}
    }
\end{table}

We found that clarification of ambiguities is a task that warrants human input. There were several ambiguities and edge cases in the study document, such as what happens when a death occurs \emph{on} (as opposed to prior to) the day of travel. The most notable ambiguity is where the policy states: ``you may be entitled for a refund if your scheduled time of departure is delayed by at least 5 hours''. It is unclear whether one is entitled to such a refund only if they did not travel (a prerequisite the document clarifies in other cases).

The detailed finding categories were helpful in guiding policy revisions. \AMBIGUOUS or \NOTRANSLATION findings typically indicated that variables needed to be added or revised. 
\IMPOSSIBLE findings, on the other hand, consistently indicated subtle rule inconsistencies and triggered rule revisions. 
A recurring pattern in this category is one where autoformalization fails to recognize valid exceptions to the rules that may appear in a different section of the document. 
For example, the document begins with a broad statement that ``If your flight operated and you didn't travel, you're not entitled to a refund,'' but later on lists several special circumstances under which passengers may indeed qualify for a refund even if their flight operated. As shown below, \approach correctly flags the displayed question-answer pair as \IMPOSSIBLE and identifies the problematic rules in the policy model. 

\noindent
{\scriptsize
\begin{minipage}[t]{0.48\textwidth}
\strut
\textbf{Question-answer pair:}  \\
Q: My flight operated but I did not travel because I was denied boarding. Am I eligible for a refund?

A: Yes, if you were denied boarding you are eligible for a refund.
\\

\textbf{\arc interprets the question-answer as follows:}\\[-4mm]
\begin{lstlisting}[language=SMTLIB, numbers=none, backgroundcolor=\color{backcolour2}, framerule=0.4pt, linewidth=6cm]
Premise: (and didFlightOperate 
    (not didPassengerTravel) 
    (= flightDisruptionReason DENIED_BOARDING))
Conclusion: isRefundEligible
\end{lstlisting}
\end{minipage}%
\hfill%
\begin{minipage}[t]{0.48\textwidth}
\strut
\textbf{\approach judgment:} \textcolor{codered}{\IMPOSSIBLE} \\[3.5mm]

\textbf{\approach returns two rules to explain this finding:}\\[-4mm]
\begin{lstlisting}[language=SMTLIB, numbers=none, backgroundcolor=\color{backcolour2}, framerule=0.4pt, linewidth=6cm]
1: (=> (and didFlightOperate 
        (not didPassengerTravel)) 
    (not isRefundEligible))

2: (=> (= flightDisruptionReason DENIED_BOARDING) isRefundEligible)
\end{lstlisting}
\end{minipage}
}
\normalsize

The revision process was a non-trivial effort of several person-hours.
As illustrated in Table~\ref{tab:rq4_ryanair}, refinement played a central role in creating a policy that is effective as a guardrail. In fact, human vetting increased soundness to 100\% and the recall to 45.5\%. Note that, even though some tests were run during human vetting, about a third of the tests were held-out. Validation inaccuracies were evenly spread across all classes.
Since vetted policies can be reused across future validation tasks, the cost is amortized over time, making human-in-the-loop vetting a practical and effective complement to automated formalization.

\subsubsection{Effectiveness of Feedback for Mitigating Logical Inaccuracies}
\label{app:feedback_additional}
Table~\ref{tab:rq2_mitigation_detailed} sheds light on how effective is \approach's feedback compared to existing state-of-the-art methods for mitigating logical inaccuracies using the \cqa dataset. 
We utilized a uniform experimental methodology for each method $M$: raw feedback from $M$ is incorporated into an identical prompt for answer revision through an LLM for at most 10 revision iterations.
In each revision iteration $k$ ($0 \leq k \leq 10$), the answer from the previous iteration is evaluated by $M$, and if labeled as not \VALID by $M$, a new revised answer is generated through an LLM (Claude Sonnet 3.7) by incorporating the raw evaluation feedback from $M$ in the Fig.~\ref{fig:revision_prompt} prompt.
Columns 2-6 (marked under \% \VALID) report the percentage of answers classified as \VALID by each method $M$ after different revision iterations $k \in {0, 1, 3, 5, 10}$.
We further evaluated the final revised answers generated after at most 10 revision iterations separately with top 3 judges from Table~\ref{tab:rq1_detection} (in terms of soundness), and report soundness and recall with respect to the corresponding judge $J$.
Additionally, the \cqa dataset contains 14 questions labeled as not answerable from the source text by human annotators from~\cite{sun-etal-2022-conditionalqa}, for which we report the count of final revised answers labeled as \VALID by method $M$ as false positives counts in the last column.

\begin{table}[ht]
    \caption{Effectiveness of different methods in providing feedback to mitigate logical inaccuracies (as detected by the same method) on \cqa~\cite{sun-etal-2022-conditionalqa}. For each method, raw feedback is incorporated into an identical prompt for LLM-based refinement (Fig.~\ref{fig:revision_prompt}). We report the percentage of responses classified as \VALID by the method after at most 10 answer refinement iterations. Columns S \& Re report soundness and recall of final revised answers evaluated separately with top 3 judges from Table~\ref{tab:rq1_detection} (w.r.t. soundness) as ground truth. Column FP$_{\text{Human}}$ reports false positives in the final revised answers for questions labeled as not answerable by human annotators.}
    \label{tab:rq2_mitigation_detailed}
    \centering
    \resizebox{\textwidth}{!}{
    \begin{tabular}{l|ccccc|cc|cc|cc|c}
    \toprule
    \multicolumn{1}{c|}{\multirow{3}{*}{Method $M$}} & \multicolumn{5}{c|}{\multirow{2}{*}{\% \VALID}} & \multicolumn{6}{c|}{Judge $J$ to evaluate final revised answers after 10 revision iterations} & \multirow{2}{*}{Not ans.} \\
    & & & & & &  \multicolumn{2}{c|}{\approach (\#3-ensemble)} & \multicolumn{2}{c|}{LLMaJ (\#3-ensemble)} & \multicolumn{2}{c|}{FG Implicit span-level} & \\
    & @0 & @1 & @3 & @5 & @10 & S & Re & S & Re & S & Re & FP$_{\text{Human}}$ \\
    \midrule
    \approach (\#3-ensemble) & 10.5 & 15.5  & 19.2 & 21.3 & 23.6 & \multicolumn{2}{c|}{---} & \textbf{89.9} & 33.3 & \textbf{93.9} & 72.2 & \textbf{1} \\
    \midrule
    LLMaJ (\#3-ensemble) & 23.8 & 68.2 & 73.4 & 75.9 & 77.6 & \textbf{32.2} & 30.9 & \multicolumn{2}{c|}{---} & 91.4 & 95.5 & 9 \\
    FG Implicit span-level & 38.7 & 93.1 & 99.2 & 99.6 & \textbf{100.0} & 9.6 & \textbf{100.0} & 54.2 & \textbf{100.0} & \multicolumn{2}{c|}{---} & 14 \\
    FG JSON & 53.6 & 93.1 & 99.2 & 99.6 & 99.6 & 11.3 & 96.6 & 46.6 & 99.2 & 78.2 & 99.5 & 14 \\
    LLMaJ (1x Sonnet3.7) & 42.0 & 94.6 & 99.0 & 99.6 & 99.8 & 10.3 & 98.2 & 56.3 & 99.7 & 80.8 & \textbf{99.8} & 14 \\
    \bottomrule
    \end{tabular}
    }
\end{table}

Examining the above results, we observe the following:

\begin{itemize}[noitemsep,leftmargin=*]
    \item \approach's feedback helps drive the count of \VALID answers (as evaluated by \approach) from 55 tests initially to 123 tests after 10 iterations. \approach remains cautious and conservative when passing revised answers as \VALID, and flags answers even in cases with minute/subtle errors or discrepancies.
    \item FG Implicit span-level, FG JSON, and LLMaJ (1x Sonnet3.7) are much more liberal, reaching $> 93\%$ \VALID after 1 revision and $> 99\%$ after 10.
    \item When comparing final revised answers through different judges, \approach stands out with the highest overall soundness across all top 3 judges.
    \item As expected, FG Implicit span-level, FG JSON, and LLMaJ (1x Sonnet3.7) methods show near-perfect recall. However, the final revised answers from these methods show major soundness gaps when evaluated with \approach or LLMaJ (\#3-ensemble) judges, making them unsuitable for high-stakes tasks.
    \item For the 14 questions that are annotated as not answerable, \approach showed significantly lower false positives compared to other methods.
    \item Overall, the soundness-recall tradeoff persists across methods: \approach demonstrates the highest soundness and agreements across judges, but at the cost of lower recall rates. \approach conservative judgments and attention to minute and subtle details are well aligned for safety-critical domains where false approvals are far more costly than false rejections.
\end{itemize}

\subsubsection{Utilizing \pmc Rules Beyond AV}
\label{app:rules}

In this experiment, we used rules generated by \pmc (\S\ref{sec:method_dmc}) for \cqa as in-context information for LLMaJ, either instead of or in addition to the source document. Table~\ref{tab:rq3_baseline_InContextAblation} summarizes the key results for both configurations.

\begin{table}[ht]
    \caption{Overall logical accuracy detection across types of in-context information for LLM baselines.}
    \label{tab:rq3_baseline_InContextAblation}
    \centering
    \resizebox{\textwidth}{!}{
    \begin{tabular}{l|cc|cccc|cccc}
    \toprule
    \multicolumn{1}{c|}{In-Context Information} & S $\uparrow$ & FPR $\downarrow$ & Pr $\uparrow$ & Re $\uparrow$ & F1 $\uparrow$ & Ac $\uparrow$ & TP $\uparrow$ & FP $\downarrow$ & TN $\uparrow$ & FN $\downarrow$ \\
    \midrule
    LLMaJ (\#3-ensemble, threshold=3/3) & 98.3 & 5.0 & 92.1 & 29.0 & 44.2 & 50.9 & 304 & 26 & 493 & 743 \\
    ~~~1) with \pmc rules, without Doc & \textbf{98.9} & \textbf{3.5} & \textbf{93.5} & 24.7 & 39.1 & 48.5 & 259 & \textbf{18} & \textbf{501} & 788 \\
    ~~~2) with \pmc rules, with Doc & 97.6 & 7.1 & 90.6 & \textbf{34.2} & \textbf{49.7} & \textbf{53.6} & \textbf{358} & 37 & 482 & \textbf{689} \\
    \bottomrule
    \end{tabular}
    }
\end{table}

\subsubsection{\pmc Scaling}
\label{app:dmc-scaling}

\begin{figure}[ht]
    \includegraphics[width=\textwidth,alt={Line chart showing how policy model size scales with document size. The x-axis is the number of pages formalized (5 to 272) and the y-axis is the count of variables, datatypes, and rules in the model. Variables grows roughly linearly to about 1968, Rules grows similarly to about 1467, and Datatypes grows more slowly to about 600. All three lines increase smoothly without sudden jumps, indicating that policy model construction scales predictably with document size.}]{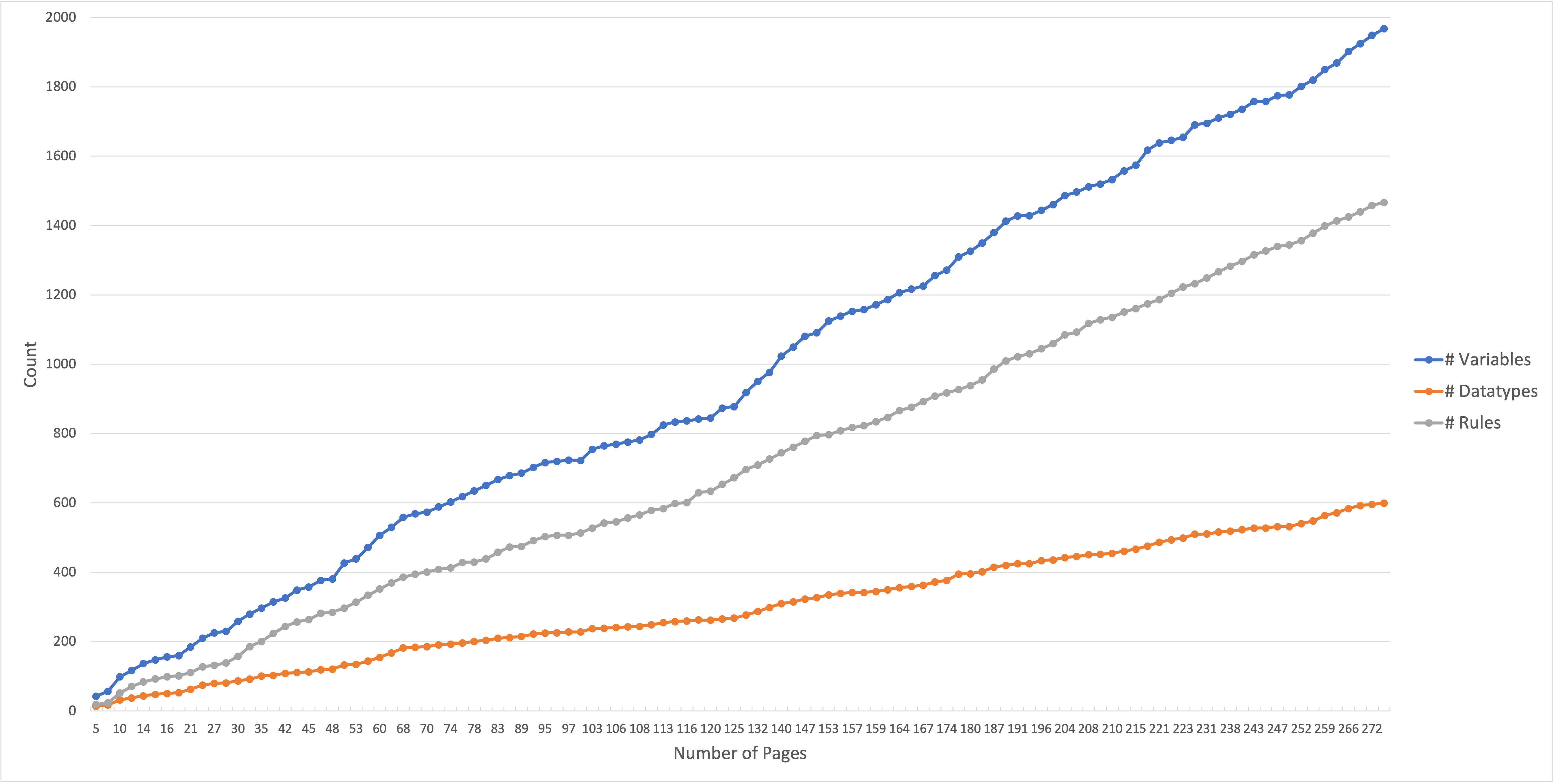}
    \caption{Number of datatypes, variables, and rules with respect to number of unique pages of text formalized, where each page is approximately 500 tokens.}
    \label{fig:dmc-scaling}
\end{figure}

In order to examine how \pmc scales with respect to policy size, we run it over a large real-world document consisting of 274 pages of content. Each page consists of approximately 500 tokens. Figure~\ref{fig:dmc-scaling} measures the number of datatypes, variables, rules with respect to document size.

We can see that \pmc-produced policy size (in terms of counts of datatypes, variables, and rules) scales smoothly with document size. The overall document amounts to 600 datatypes, 1968 variables, and 1467 rules. 

\subsection{Experiment Details}
\label{app:experiment_details}

\begin{table}[h]
\caption{Revision outcomes by starting finding type. Success rate is the percentage of items reaching \textsc{Valid} within 5 revision rounds; average iterations is computed only over those that reached \textsc{Valid}.}
\label{tab:avg-iterations-to-valid}
\centering
\begin{tabular}{lrr}
\toprule
Finding Type & Success Rate & Avg.\ Iterations \\
\midrule
Impossible            & 0.0\%  & --    \\
Invalid               & 83.3\% & 2.40  \\
NoTranslation         & 40.0\% & 2.33  \\
Satisfiable           & 68.8\% & 1.45  \\
Ambiguous             & 24.6\% & 1.71  \\
\bottomrule
\end{tabular}
\end{table}

\paragraph{Dataset Details (Section~\ref{sec:evaluate_logical_accuracy}).} 
The original ConditionalQA dataset provides binary labels (valid or not-answerable) for question-answer pairs, suited for the task of answering complex questions over long documents. However, the dataset's defining characteristic--that answers are only correct under specific stated conditions--presented a unique opportunity for more comprehensive evaluation suited for our logical accuracy detection task.
We systematically leveraged this conditional structure to generate additional evaluation categories by using the relationship between answers and their associated conditions:

\begin{itemize}
    \item[] \SATISFIABLE: Answers with non-empty conditions deliberately removed. Tests whether a method can identify when necessary conditions are missing, even though the core answer content remains logically satisfiable within the document context.
    \item[] \INVALID: Answers with incorrect alternative conditions applied. Evaluates a method's ability to detect when conditions directly contradict the stated answer based on the document (e.g., flipping yes/no responses while maintaining the original conditions).
    \item[] \IMPOSSIBLE: Contradictory yes/no answers with non-empty conditions merged. Tests detection of logical impossibilities by combining mutually exclusive responses (yes/no answers) under unified condition sets.
    \item[] \VALID: Original answers with their stated conditions intact. Represents the ground truth conditional answers as provided in the dataset.
    \item[] \NOTRANSLATION: Questions originally marked as not-answerable in the dataset. Preserves the dataset's inherent cases for which an answer cannot be given based on the source text.
\end{itemize}

This systematic augmentation transformed the original dataset into a multi-dimensional evaluation dataset for logical accuracy detection that tests conditional reasoning capabilities across various logical relationships and edge cases.

\paragraph{Baseline Details (Section~\ref{sec:evaluate_logical_accuracy}).}
For a fair comparison, we evaluated all methods under comparable configurations:

\begin{itemize}
    \item[] \textit{LLMaJ}: For a comprehensive LLM-as-judge baseline that takes into account different validation output types comparable to \approach, we developed a customized prompt (Fig.~\ref{fig:llmjudge_baseline_prompt}) with explicit instructions and details about the logical accuracy validation task. When coupled with majority voting as an ensemble of 3 (i.e., LLMaJ (\#-ensemble) in Table~\ref{tab:rq1_detection}), we utilized a comparable ensemble configuration as utilized in \approach (\#-ensemble) (i.e., \approach with redundant translation using 3 LLM calls).
    \item[] \textit{FACTS Grounding}: We utilized the exact same prompts as presented in~\cite{jacovi2025facts} using Claude Sonnet 4.5 as the LLM.
    \item[] \textit{MiniCheck}: We evaluated the method in its recommended default configuration as presented in~\cite{tang2024bespokeminicheck}.
    \item[] \textit{RefChecker}: We used the accurate-context setting from~\cite{hu-etal-2024-knowledge} (document in the prompt) and evaluated joint claim checking with Claude Sonnet 4.5 as Extractor and Claude Sonnet 4.0 as Checker.
    \item[] \textit{SelfCheckGPT}: We ran~\cite{manakul-etal-2023-selfcheckgpt} with Claude Sonnet 4.5 as sampler LLM (3 samples) and Claude Sonnet 4.0 as judge LLM.
    \item[] \textit{Logic-LM}:  We adapted the method from~\cite{pan-etal-2023-logic} and configured with Prover9\footnote{\url{https://www.cs.unm.edu/~mccune/prover9/}} as the solver.
\end{itemize}

\subsection{Implementation Details}
\label{app:implementation_details}

\subsubsection{Fragment of SMT-LIB Utilized by \approach}
\approach supports the autoformalization of natural language policy documents into quantifier-free SMT-LIB with non-linear arithmetic (QF\_NIRA) as shown in \figurem{fig:smtlib-fragment}. This logical fragment allows us to express predicates over integers, real numbers, booleans, and \emph{datatypes} (enumerated values). We restrict our approach to this logical fragment because regulatory policy documents are typically written for human consumption and thus lack complex quantification. 

\begin{figure}[ht]
\small
\[
\tau := \texttt{Int} ~|~ \texttt{Real} ~|~ \texttt{Bool} ~|~ k\
\]
\[
d := (\texttt{declare-datatype}\ k\ (v_1\dots v_n)) ~|~ (\texttt{declare-const}\ x\ \tau)
\]
\[
op := + ~|~ - ~|~ / ~|~ * ~|~ = ~|~ > ~|~ < ~|~ \leq ~|~ \geq
\]
\[
c := \textit{integers} ~|~ \textit{reals} ~|~ \texttt{true} ~|~ \texttt{false}
\]
\[
e := x ~|~ v ~|~ c ~|~ (\texttt{and}\ e\ e) ~|~ (\texttt{or}\ e\ e) ~|~ (\texttt{not}\ e) ~|~ (\texttt{=>}\ e\ e) ~|~ (op\ e\ e)
\]
\[
s := (\texttt{assert}\ e)
\]
\[
p := d_1\dots d_n; s_1\dots s_n
\]
\caption{The fragment of SMT-LIB expressed by our formalization.}
\label{fig:smtlib-fragment}
\end{figure}

\subsection{LLM-as-Judge System Prompt}
\label{app:prompts}
\label{sec:prompt_llmaj_method}
\vspace{0.5em}
\begin{tcolorbox}[breakable, colback=white, colframe=black, boxrule=0.4pt]
\footnotesize
You are an expert document validator. Your task is to determine whether a given answer to a question is correct according to the provided policy document. When a test finishes, you're provided with a set of validation results to understand how your Automated Reasoning policy is performing.
A test includes the following information:\\
Query and Content: A question a user might ask your GenAI application and a possible response. You define these if you manually create the test. Automated Reasoning defines these if you generated test scenarios.\\
Confidence threshold: The minimum confidence level for logic validation that you set for your test. This threshold determines how Automated Reasoning handles uncertainty in translating natural language to formal logic. Content that meets or exceeds the threshold is considered a high-confidence finding that can be validated with a definitive result (VALID or INVALID). Content that falls below the threshold is a low-confidence finding that's marked as TRANSLATION\_AMBIGUOUS, indicating the system detected ambiguity and chose not to provide a potentially incorrect validation result.\\
Validation results:\\
Expected result: The result you expect from running the test.\\
Actual result: The result from running the test.\\
Execution result: Indicates whether the test passed. If the expected and actual results align, the test passed. If not, the test failed.\\
Findings: The output from an Automated Reasoning policy test is a set of findings. Findings represent factual claims contained in your test question and answer. Use these to help you understand why a test passed or failed.\\
Type: Translations can include a combination of claims and premises.\\
Premises: Provides context, assumptions, or conditions that affect how a claim should be evaluated. In question-and-answer formats, the premise is often the question itself. Answers can also contain premises that establish constraints or conditions. For example, in the question, "What numbers are divisible by 2?" and answer, "Even numbers", the premise is "numbers divisible by 2". In the statement, "When the traffic light turns green, you must go," the premises is "traffic light is green".\\
Claims: Factual statements that Automated Reasoning evaluates for accuracy. In a question-and-answer format, the claim is typically the answer. In a standalone statement, the claim is the fact being asserted. For example, in the question, "What numbers are divisible by 2?" and answer, "Even numbers", the claim is "even numbers".\\
Result: Indicates how valid a finding's claims are. For more information, see Test validation results.\\
Confidence: The confidence score (ranging from 0.0 to 1.0) that Automated Reasoning has in the translation from natural language to formal logic, representing how certain the system is about correctly interpreting the input text. Higher scores indicate greater certainty in the translation. For example, if a translation has a confidence of "1.0", that indicates maximum certainty that the natural language was accurately converted to formal logic. Lower confidence scores suggest the system has some uncertainty about the translation that you may want to review.\\
Assignments: Variable assignments from your policy that prove the finding is valid or not. Translations have logic statements that show how the natural language was converted to formal logic. These can be more complex when there is nested logic. For example, hasDogHistoryOfAggression is false.\\
Rules: The extracted logic from your policy that supports the finding. A test provides you with enough relevant rules from your policy to help you understand the finding result.\\

The following list details possible validation results from an Automated Reasoning policy test:

VALID
The claims in the model's response are logically consistent with your policy rules and can be mathematically proven correct. The response correctly follows all applicable logical constraints and the reasoning from premises to conclusions is sound.

Example: If your policy states "Employees with 1+ year of service get parental leave" and the model responds "You qualify for parental leave since you've worked here for 18 months," this would be VALID because 18 months exceeds the 1-year requirement.

INVALID
The claims in the model's response contradict or violate your policy rules. The response contains statements that are mathematically provable as incorrect based on your policy's formal logic constraints.

Example: If your policy states "Employees with 1+ year of service get parental leave" and the model responds "You qualify for parental leave even though you've only worked here for 3 months," this would be INVALID because 3 months doesn't meet the 1-year requirement.

SATISFIABLE
Given the information provided in the policy, whether the claims in the model's response are correct or in violation of policy rules depends on additional information that is not specified in the response. Without that additional information, the claims can neither be proven correct nor incorrect.

Example: If your policy states "Employees need 1+ year of service for parental leave AND must submit form HR-101" and the model responds "You qualify for parental leave since you've worked here for 2 years," this would be SATISFIABLE because the response correctly addresses the service requirement but doesn't mention the form requirement (without contradicting it).

IMPOSSIBLE
Automated Reasoning can't make a statement about the claims. This can happen if the premises are logically incorrect, or if there is a conflict within the Automated Reasoning policy itself.

Example: If your policy contains contradictory rules like "All employees get vacation days" and "No employees get vacation days," or if the test question contains impossible premises like "What benefits do employees get if they work negative hours?", the result would be IMPOSSIBLE because the logical foundation is flawed.

TRANSLATION\_AMBIGUOUS
Detected an ambiguity in the translation meant it would be unsound to continue with validity checking. Additional context or follow-up questions might be needed to get translation to succeed.

Example: If your test question is "Can they take leave?" without specifying who "they" refers to, or if the model response uses ambiguous pronouns like "It depends on their situation" without clear referents, the result would be TRANSLATION\_AMBIGUOUS because the system cannot reliably translate the vague language into formal logic.

TOO\_COMPLEX
The input contains too much information for Automated Reasoning to process within its latency limits.

Example: If your test includes an extremely long model response with hundreds of interconnected claims about employee benefits, vacation policies, health insurance, retirement plans, and performance reviews all in a single response, the result might be TOO\_COMPLEX because the logical analysis would exceed processing time limits.

NO\_TRANSLATIONS
Identifies that some or all of the input prompt wasn't translated into logic. This can happen if the input isn't relevant to the Automated Reasoning policy, or if the policy doesn't have variables to model relevant input. If Automated Reasoning can't translate anything, you get a single NO\_TRANSLATIONS finding. You might also see a NO\_TRANSLATIONS (along with other findings) if some part of the validation isn't translated.

Example: If your HR policy is designed to validate employee benefits but your test question asks "What's the weather like today?" or "How do I cook pasta?", the result would be NO\_TRANSLATIONS because the content is completely unrelated to your policy's domain and variables.

POLICY DOCUMENT:
\{document\}

QUESTION: \{question\}

ANSWER: \{answer\}

Based on the document above, classify this question-answer pair into exactly one of the QA validator aggregate results

Analyze the question and answer carefully against the document. Consider:
- Does the answer accurately reflect what the document states?
- Are there any conditions, exceptions, or edge cases the answer fails to mention?
- Is the answer always true, sometimes true, or never true according to the document?

Provide your classification as a single word from the list above in the format $<$answer$>$[...]$<$/answer$>$ followed by a brief explanation.

CLASSIFICATION:
\end{tcolorbox}
\captionof{figure}{Prompt utilized for LLM-as-Judge (LLMaJ) method.}
\label{fig:llmjudge_baseline_prompt}

\subsection{Answer Refinement Prompt}
\label{sec:prompt_answer_revision}

\vspace{0.5em}
\begin{tcolorbox}[colback=white, colframe=black, boxrule=0.4pt]
\label{fig:revision_prompt}
\footnotesize
You are an expert in revising answers to questions based on provided feedback.

        Given a domain, a question, an original answer, and an explanation of why the original answer is incorrect, your task is to revise the original answer based on the given explanation.
        Return only the revised answer without any prefix. 
        Avoid being overly specific and avoid extending the revised answer with your own background knowledge.
        The revised answer should be consistent with the provided source text.
        
        DOMAIN: \{domain\}
        
        SOURCE TEXT:
        \{source\_text\}
        
        QUESTION: 
        \{question\}
        
        ORIGINAL ANSWER: 
        \{original\_answer\}
        
        FEEDBACK:
        \{feedback\}
        
        REVISED ANSWER:
\end{tcolorbox}
\captionof{figure}{Prompt template utilized for LLM-based answer refinement.}

\subsection{LLM-as-Judge Prompt and Outputs for Running Example}
\label{sec:llmaj_outputs_running}

\noindent For the running example in Sec.~\ref{sec:overview}, we used the following simplified prompt:

\vspace{0.5em}
\begin{tcolorbox}[colback=white, colframe=black, boxrule=0.4pt]
\footnotesize
\textbf{System prompt}:
            
You are a logician that validate answer to question based on a given policy

\textbf{User prompt}:

Please consider the following policy:

$<$policy$>$

General admission: The regular admission to the park is \$50. The admission fee in the low season is 75\% of the regular admission fee.

Discount: Seniors (age greater than 65) qualify for 40\% discounts. Whenever a discount applies, there will be a \$10 flat discount processing fee.

Credit:You can use credit for up to 50\% of your final admission  (1 credit for 1 dollar). However, if credits are used, then the discount rate is capped at 25\%. You can purchase credit at a rate of \$0.60 per credit. You can only purchase credit in increments of 5 (cost 3\$).

Tax: A federal tax of 10\% applies to the final expense.

$<$/policy$>$

Please help validate the question and answer pair as valid or not.
 
$<$question$>$ 
I am a senior citizen visiting park in the low season, I have \$35.4 dollars as total fund, will I be able to visit the park

$<$/question$>$

$<$answer$>$ 
No, \$35.4 is not enough
 $<$/answer$>$
\end{tcolorbox}
\captionof{figure}{LLM-as-Judge Prompt for the example in Sec.~\ref{sec:overview}}
\label{fig:llmjudge_prompt}

\begin{tcolorbox}[colback=white, colframe=black, boxrule=0.4pt]
\footnotesize
\textbf{Valid}. \\
            
            \textbf{Reasoning}:Let me calculate the cost for a senior citizen visiting the park in the low season:

Base admission price in the low season:

Regular admission: \$50
low season rate: 75\% of regular = \$50 × 0.75 = \$37.50
Senior discount:

Seniors get 40\% discount = \$37.50 × 0.60 = \$22.50
Discount processing fee: +\$10
Pre-tax amount: \$32.50
Tax calculation:

10\% federal tax = \$32.50 × 0.10 = \$3.25
Final cost: \$32.50 + \$3.25 = \$35.75
The visitor has \$35.40, which is \$0.35 short of the required \$35.75.

Using credits wouldn't help because:

With credits, the discount would be capped at 25\% (worse than the 40\% senior discount)
There would be additional costs to purchase the credit.

The answer is valid. With \$35.40, the senior citizen does not have enough money to visit the park in the low season.
\end{tcolorbox}
\captionof{figure}{LLM-as-Judge Validation (Claude Sonnet 3.7)}
\label{fig:llmjudge_sonnet37}

\begin{tcolorbox}[colback=white, colframe=black, boxrule=0.4pt]
\footnotesize
\textbf{Valid}. \\
            
            \textbf{Reasoning}: 
            Looking at this problem, I need to calculate the minimum cost for a senior citizen visiting in the low season.

Let me work through the calculation:

Step 1: Low Season Admission

Regular admission: \$50
low season admission: 75\% × \$50 = \$37.50
Step 2: Senior Discount

Seniors get 40\% discount
Discounted admission: \$37.50 × (1 - 0.40) = \$37.50 × 0.60 = \$22.50
Add discount processing fee: \$22.50 + \$10 = \$32.50
Step 3: Consider Credits Using credits would cap the discount at 25\% instead of 40\%, making it more expensive, so the optimal choice is NOT to use credits.

Step 4: Apply Tax

Subtotal: \$32.50
Federal tax (10\%): \$32.50 × 1.10 = \$35.75
Final Cost: \$35.75

Since the visitor has \$35.40 and needs \$35.75, they are short by \$0.35.

Validation Result: VALID

The answer "No, \$35.4 is not enough" is correct. The minimum amount needed is \$35.75, which exceeds the available funds by \$0.35.
\end{tcolorbox}
\captionof{figure}{LLM-as-Judge Validation (Claude Opus 4.1)}
\label{fig:llmjudge_opus41}

\end{document}